\documentclass[authoryear,review,preprint,12pt]{elsarticle}
\usepackage{setspace}
\usepackage{amssymb}
\usepackage[T1]{fontenc}

\journal{arXiv}

\begin{document}

\begin{frontmatter}



\title{Predictive modeling of microbiological seawater quality classification in karst region using cascade model}

\author[inst1,inst3]{Ivana Lu\v{c}in}
\author[inst1,inst3]{Sini\v{s}a Dru\v{z}eta}
\author[inst2,inst3]{Goran Mau\v{s}a}
\author[inst1]{Marta Alvir}
\author[inst1,inst3]{Luka Grb\v{c}i{\'c}}
\author[inst3,inst4,inst5]{Darija Vuki{\'c} Lu\v{s}i{\'c}}
\author[inst1,inst3]{Ante Sikirica}
\author{Lado Kranj\v{c}evi{\'c}\corref{cor1}\fnref{inst1,inst3}}
\cortext[cor1]{Corresponding author}
\ead{lado.kranjcevic@riteh.hr}

\affiliation[inst1]{organization={Department of Fluid Mechanics and Computational Engineering, Faculty of Engineering, University of Rijeka},
            addressline={Vukovarska 58}, 
            city={Rijeka},
            postcode={51000}, 
            country={Croatia}}

\affiliation[inst2]{organization={Department of Computer Engineering, Faculty of Engineering, University of Rijeka},
            addressline={Vukovarska 58}, 
            city={Rijeka},
            postcode={51000}, 
            country={Croatia}}

\affiliation[inst3]{organization={Center for Advanced Computing and Modelling, University of Rijeka},
            addressline={Radmile Matej\v{c}i{\'c} 2}, 
            city={Rijeka},
            postcode={51000}, 
            country={Croatia}}

\affiliation[inst4]{organization={Department of Environmental Health, Faculty of Medicine, University of Rijeka},
            addressline={Bra{\'c}e Branchetta 20/1}, 
            city={Rijeka},
            postcode={51000}, 
            country={Croatia}}

\affiliation[inst5]{organization={Department of Environmental Health, Teaching Institute of Public Health of Primorje-Gorski Kotar County},
            addressline={Kre\v{s}imirova 52a}, 
            city={Rijeka},
            postcode={51000}, 
            country={Croatia}}            

\begin{abstract}
In this paper, an in-depth analysis of \textit{Escherichia coli} seawater measurements during the bathing season in the city of Rijeka, Croatia was conducted.  Submerged sources of groundwater were observed at several measurement locations which could be the cause for increased \textit{E. coli} values. This specificity of karst terrain is usually not considered during the monitoring process, thus a novel measurement methodology is proposed. A cascade machine learning model is used to predict coastal water quality based on meteorological data, which improves the level of accuracy due to data imbalance resulting from rare occurrences of measurements with reduced water quality. Currently, the cascade model is employed as a filter method, where measurements not classified as excellent quality need to be further analyzed. However, with improvements proposed in the paper, the cascade model could be ultimately used as a standalone method.
\end{abstract}




\end{frontmatter}



\section{Introduction}


Microbiological contamination presents a great concern in areas where water bodies are used for recreational activities since the existence of pathogens can cause serious health problems \citep{solo2016beach}. This is especially important for tourism-oriented countries, such as Croatia, since bathing locations with high water quality can attract tourists, and maintenance of such favourable repute is one of the main priorities. Currently, the main methodology for water quality classification consists of fortnightly measurements of feacal indicator bacteria, such as \textit{Escherichia Coli} or enterococci. Unfortunately, measurements are temporary and spatially sparse as they are expensive and time consumable. This is a considerable problem since studies observed that number of microbes has a high temporal and spatial variation \citep{ekklesia2015temporal,luvsic2017temporal}. Additionally, currently in Croatia, sampling and laboratory testing take about 2.5 days, thus the information is already outdated by the time it is obtained. Therefore, methods for predicting water quality integrating meteorological data are increasingly being investigated.

Many factors need to be taken into consideration when investigating microbe concentrations in the seawater, such as solar radiation, tides, wind intensity and direction, rainfall, the density of bathers, presence of rivers and canals near the bathing area, etc. \citep{profiles2009best,he2008water, cho2010meteorological}. In a number of studies, rainfall was deemed greatly influential for microbiological contamination both in coastal \citep{dwight2011influence,luvsic2017temporal,he2019storm} and underground waters \citep{knierim2015quantifying,mance2018environmental,buckerfield2019rainfall}. A more detailed overview of numerical modeling approaches and the influence of normal and extreme storm events on \textit{E. coli} values in coastal waters was reviewed in \citet{weiskerger2020numerical}.

When investigating the water quality of coastal areas, specifics of each location must be taken into consideration. \citet{he2019storm} conducted an analysis of two beaches, approximately 20 km apart in China, after a single storm event where a difference in water quality was observed due to distinct beach environments. \citet{viau2011bacterial} investigated bacterial pathogens in Hawaiian coastal streams which were shown to be pollution sources associated with beach locations. In \citet{kucuksezgin2019assessment} the enclosed bay of Izmir Bay, Turkey was analyzed, where domestic and industrial wastes contribute to reduced water quality. In \citet{verga2020assessment} an analysis of seasonal and spatial variability of water quality in Patagonia, Argentina was conducted. Observed problem with sewage and draining systems was linked to insufficient dilution of wastewater in seawater. \citet{chahouri2021combined} conducted an assessment of both beach and estuary location in Agadir bay, South-West Morocco. It is noted that estuary location is influenced by effluents from untreated wastewater and agricultural activities and the tourist beach is center of human activities, where greater fecal streptococci loads were observed at the estuary location. These studies indicate need for consideration of both urban development and geographical specifics of each location.

The terrain of the Croatian coast is mostly of karst type and characterized by high porosity and numerous subterranean channels, which makes it very vulnerable to pollution since surface water can quickly and easily enter groundwater \citep{pikelj2013eastern}. Investigation of \textit{E. coli} pollution in karst was conducted in a number of studies \citep{davis2005escherichia,laroche2010transport}. Sources of groundwater contamination can include landfills \citep{kogovvsek2013increase}, sewage outflows \citep{heinz2009water, stange2020occurrence}, agricultural or urban land-use type \citep{reed2011differences, buckerfield2019rainfall}, etc. For this reason, the water quality of karst aquifers used for drinking water is regularly monitored with special care, while karst aquifers not used for drinking water are not regularly monitored due to their reduced importance.

With growing interest in increased protection of water surfaces, prediction modeling is being used to provide information to the general public regarding potential health risks. \citet{he2008water} used Artificial Neural Network (ANN) to predict water quality regarding stormwater runoff with reported less than 10\% false positive or negative rates. In \citet{de2018optimising} and \citet{de2018developing} regression models for predicting fecal indicators in coastal waters are developed and optimized, where cumulative solar radiation and cumulative rainfall values were shown to greatly influence the prediction of fecal pollution. \citet{he2019storm} used Multiple linear regression (MLR) model to predict pathogen contamination using environmental data collected during the storm event. \citet{grbvcic2021coastal} investigated the efficiency of different machine learning algorithms to
predict E. Coli and enterococci values based on environmental features.

The main premise of the proposed work is that a prediction model can be created which would use meteorological data for predicting \textit{E. coli} values, by use of which forecasts could be made and warnings to the general public can be given in advance. The main contribution to this goal was made by utilising Random Forest classifier to predict water quality based on meteorological data. Data used for model training is obtained from available measurements of water quality during the bathing seasons 2009-2020 for Rijeka, Croatia. An investigation of different clustering methods of bathing locations was conducted with the addition of feature analysis. Additionally, a novel cascade prediction model framework, aimed at classifying measurements as excellent water quality, is proposed. Due to comprising a series of prediction models, it enables model fine-tuning for different physical processes with increased prediction accuracy. The proposed model provides great flexibility and as such can be used on pollution measurement datasets with sparse cases of high pollution, which are the most common. In-depth analysis of measured data indicated the potential influence of submerged coastal springs which are specific for karst soil. In previous research \citep{luvsic2017temporal}, these springs were not considered as possible pollution sources because they predominantly dry up during the bathing season. However, during the investigation of hydrogeological data for the monitored region, it was found that some of them are active throughout the whole year. Considering these new findings further research directions are presented in the discussion section.

\section{Materials and methods}

\subsection{Data collection}

\begin{figure}[!ht]
\includegraphics[width=\textwidth]{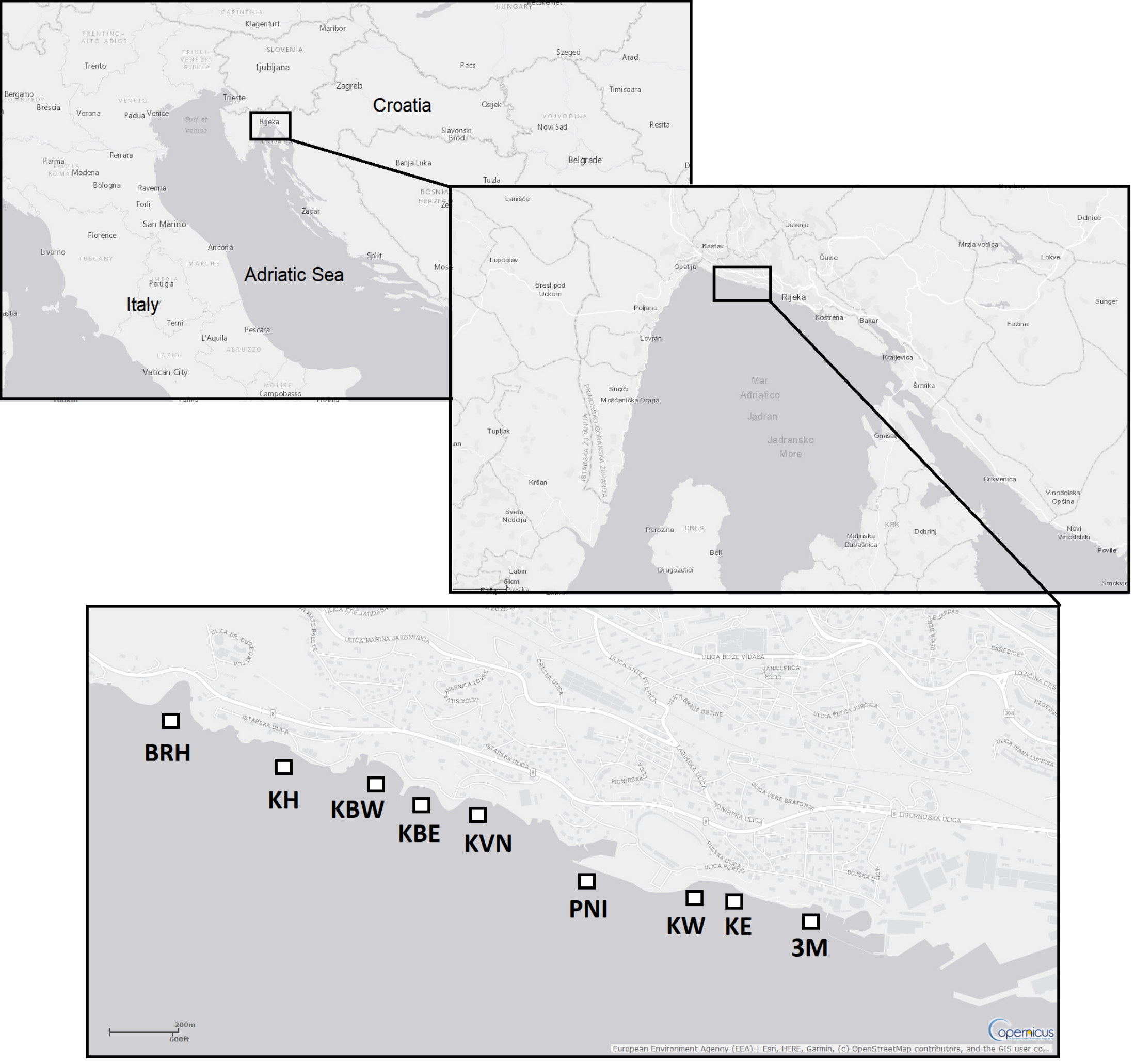}
\caption{Measurement locations in Rijeka Bay (Kvarner Bay, Croatia) \citep{copernicus}.}
\label{fig:measurement_locations}
\end{figure}

The city of Rijeka is located in Kvarner Bay and is the third-largest Croatian city with important industrial locations such as shipyard and port, but with the increasing tendency to become a recognized touristic location. The city has Mediterranean climate, with dry and warm summers, and its surrounding is also characterized by a large amount of rainfall due to the proximity of Dinaric Alps. In the city of Rijeka precipitation is estimated at about 1540 mm per year, with 550 mm per year for period May to September. The average temperature of the warmest months (July and August) is 23.1$^{\circ}$C \citep{zaninovic2008klimatski}. The bathing season usually lasts from mid-May until the end of September, and in that period regular measurements of water quality are conducted. Measurements from 9 locations on the west side of the city are considered for analysis with mentioned locations spread over roughly 2 km length of the coastline.
Locations of the measurement points can be observed in Figure \ref{fig:measurement_locations}. Regular measurements, with fortnightly intervals, from 2009 to 2020, are used as data inputs for the prediction model. Samples were taken from the boat in a short timeframe. Samples from all considered measuring points were collected, on average, within $30$ minutes. Consequently, differences in atmospheric conditions are very small or nonexistent, between measurements taken on the same day. For each measurement point, samples are taken at a similar distance (20 meters) from the coast. During several months in 2012 and 2014, additional measurements were conducted every 4 hours for 5 measuring points: KBW, KBE, KW, KE, and 3M (see Figure \ref{fig:measurement_locations}). Additional measurements taken in the morning, which is the time when regular measurements are conducted, were also included in the analyzed dataset to increase the number of measurements. Additional measurements were always taken from the same location at the coast. Water temperature, salinity, and air temperature were measured in situ and \textit{E. coli} value was analyzed in the laboratory from the collected sample.
For \textit{E. coli} enumeration membrane filtration technique was used, according to the ISO-9308-1 method for period 2009-2017 and temperature-modified ISO-9308-1 method for period 2018-2021. Cultivation was performed on CCA nutrient media (Chromogenic Coliform agar, Biolife Italiana S.r.l., Milan, Italy) for 4 h at 36 $\pm{}$ 2$^{\circ}$C followed by 20 h incubation at 44 $\pm{}$  0.5$^{\circ}$C \citep{jozic2018report,jozic2018performance}. Further details on data collection and analysis are given in \citet{luvsic2017temporal}. 

In total, considered measurements have 1137 records. It must be noted that not all measuring points have the same number of measurements, since 5 locations (KBW, KBE, KW, KE, and 3M) have additional measurements and measurements for PNI started in 2019. 

\subsection{Environmental parameters}

Features chosen for prediction modelling were meteorological data obtained during measurements: water temperature, air temperature, and salinity. Additionally, solar irradiance as current
Global Horizontal Irradiance (GHI) and cumulative irradiance for the past 4 hours are also considered. The chosen data was taken from \citet{solcastdata} database. The rainfall data was obtained from the Croatian Meteorological and Hydrological Service (DHMZ). Different combinations of cumulative rainfall values, such as previous $24$ hours, previous $48$ hours, etc., were also considered since in a number of previous studies influence of rainfall, especially storm events, were investigated \citep{he2019storm,weiskerger2020numerical}. It was observed that a very small number of measurements have any rainfall from the previous several days, thus cumulative sums from $4-7$ and $7-14$ days are considered as a possible indication of soil saturation, which can happen if a larger amount of rain is present during a longer period of time. If soil is saturated, new rain can influence the activation of underground sources in the sea, which can increase the amount of \textit{E. coli}.

\subsection{Data analysis and preparation}

\begin{figure}[!ht]
\includegraphics[width=\textwidth]{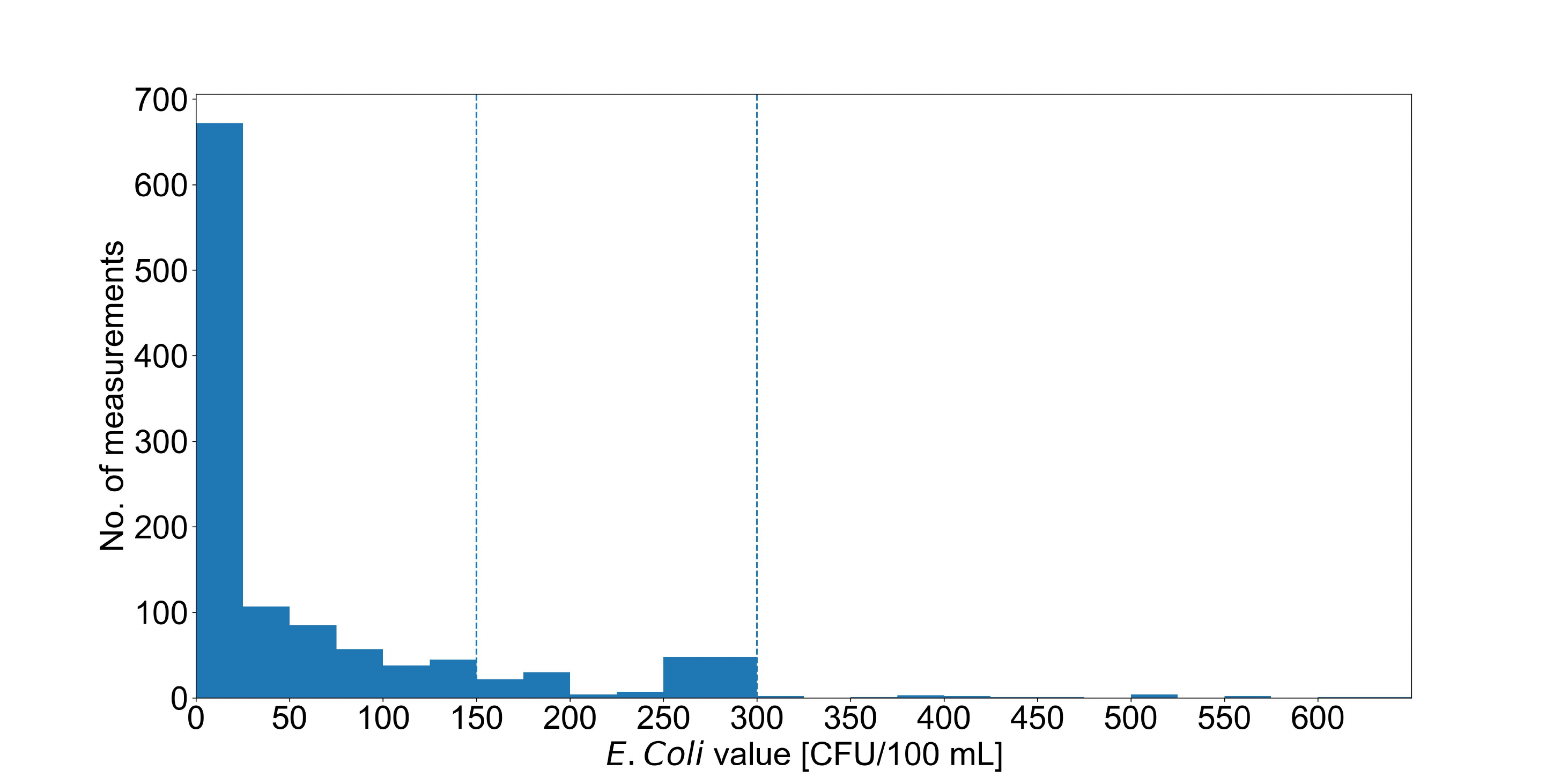}
\caption{Histogram of \textit{E. coli} measurements. Vertical lines indicate limits for excellent and sufficient water quality.}
\label{fig:histogram_measurements}
\end{figure}

The considered measurement points are chosen for prediction modeling since, historically, these locations have lower water quality than other bathing locations in the city, even though only a small amount of these measurements show less than excellent water quality. EU legislation \citep{ec2006european} prescribe two fecal indicator bacteria, \textit{E. coli} and enterococci where in \citet{luvsic2017temporal} it was observed that \textit{E. coli} criteria in Croatian legislation \citep{cro2008directive} are more stringent. Thus in this work, the prediction models are investigated considering only \textit{E. coli} limits. Bathing water quality can be divided into three categories, where, by Croatian standards, excellent water quality is for \textit{E. coli} measurements in the range 0-150 CFU/100 mL (in at least 95\% of samples), good water quality in \textit{E. coli} range 150-300 CFU/100 mL in at least 95\% of samples, and sufficient quality for the range up to 300 CFU/100 mL in at least 90\% of samples. EU criteria allow \textit{E. coli} values up to 250 CFU/100 mL (in at least 95\% of samples) for excellent water quality, good water quality in the range 250-500 CFU/100 mL (in at least 95\% of samples) and sufficient water quality for \textit{E. coli} values up to 500 CFU/100 mL in at least 90\% of samples. If Croatian criteria is applied, and value for \textit{E. coli} contamination is taken as $300$ CFU/100 mL, there are $21$ records in the dataset with \textit{E. coli} values above that threshold. If contamination measure is taken as $150$ CFU/100 mL, $122$ records have \textit{E. coli} value above that threshold. It can be observed that there is only $11$\% of less-than-excellent and only $1.8$\% of less-than-sufficient water quality measurements. Thus, an in-depth analysis of these cases is performed. Histogram of considered measurements can be observed in the Figure \ref{fig:histogram_measurements}.

It was observed that greater values of \textit{E. coli} can be roughly grouped in three types of pollution events. The first are unexpected and unexplained surges in \textit{E. coli} concentration during the monitoring. They are followed by control measurements and are considered as an accidental event. These measurements are considered outliers since they show no correlation with meteorological data, thus it is reasonable to believe that these are singular incident events. Therefore these measurements are removed from the data set to improve the prediction model performance. After this correction, $1133$ measurement remained, with $17$ records having \textit{E. coli} value above $300$ CFU/100 mL ($1.5$\% of measurements) and $118$ records above $150$ CFU/100 mL ($10.4$\% of measurements). The second group of events are typically occurring during spring, where it was observed that water salinity is reduced in all measuring points, which can be explained by the greater amount of precipitation during the spring period which could lead to greater \textit{E. coli} concentrations. The average salinity value for the period from May to mid-June was $30.6$ with an average \textit{E. coli} value $72.3$ CFU/100 mL, and for the period from mid-June to September, average salinity was $34$ with an average \textit{E. coli} value $46$ CFU/100 mL, which supports the presented assumption. The third type of events are the occurrences where it was observed that some measuring points show higher \textit{E. coli} value coupled with lower water temperature and lower water salinity than other nearby measuring points, during the same measurement period. This indicates the presence of local sources of groundwater at these measuring points. It must be noted that not all measurements with lower water quality can be put in these categories, thus prediction modelling needs to be utilized to find additional correlations.

The mean value of salinity was analysed for all measuring points to investigate the possible correlation with \textit{E. coli} value and results are presented in Table \ref{table:beach_analysis}. It can be observed that locations with greater average \textit{E. coli} value also have smaller average salinity. This can be explained by the fact that underground water sources, which presence can be observed through reduced water salinity, collect bacteria from the watershed area and transport them into the seawater. However, this assumption needs to be further investigated.

\begin{table}[!ht]
\centering
\caption{Salinity and \textit{E. coli} characteristics of measuring points.}
 \begin{tabular}{ c c c c c }
 \hline
{Measuring point} & \multicolumn{2}{c}{\textit{E. coli} (CFU/100 mL)} & \multicolumn{2}{c}{Salinity}\\
{(number of measurements)} & {Mean} & {Median} & {Mean} & {Median} \\
\hline
{BRH (122)} & {36} & {5} & {35} & {36} \\
{KH (123)} & {47} & {7} & {34.5} & {35.7} \\
{KBW (144)} & {36} & {7} & {34.7} & {35.9} \\
{KBE (149)} & {26} & {5} & {34.8} & {36} \\
{KVN (119)} & {35.8} & {8} & {34.4} & {35.5} \\
{PNI (20)} & {56.8} & {26} & {31.8} & {34.5} \\
{KW (151)} & {78} & {35} & {31.6} & {33.3} \\
{KE (155)} & {86.4} & {60} & {30} & {32.2} \\
{3M (150)} & {72.3} & {25} & {30.5} & {32.9} \\
\hline

 \end{tabular}
 \label{table:beach_analysis}
\end{table}

\subsection{Random Forest classifier}

Machine learning algorithms are designed to find an underlying correlation or patterns between the data input and data output to provide a prediction for unseen data. Machine learning algorithms can be divided into regression and classification algorithms, where the first group of algorithms aims to predict the exact value of the output variable, while the other try to separate data into predefined groups. Since the problem considered in this paper, by nature of corresponding regulation, deals with water quality groups, a classifier algorithm was considered for the prediction of water quality. Prediction models were constructed with only two classes, i.e. a prediction is made whether \textit{E. coli} value is above or below the considered limit. Random Forest classifier implementation in the Python library Scikit-learn \citep{pedregosa2011scikit} version 0.20.3 was used.

Random Forest classifier is an ensemble type of machine learning algorithm which was first proposed by \citet{breiman2001random}. It consists of multiple decision trees which stand as independent prediction models. The bootstrap method is used to provide a unique subset for the training of each decision tree while the aggregation method is used to count the class with the most prediction occurrences which is then considered as the prediction of the Random Forest model.

Different combinations of Random Forest Classifier parameters were investigated, where best results were obtained for $100$ estimators, maximum depth of $10$, and the minimum number of samples required to split an internal node equal to $6$. All other parameters were kept at default values. Considered parameters were obtained for prediction model trained on the first group of uniformly distributed data with \textit{E. coli} classification limit $150$ CFU/100 mL which is Croatian criteria for excellent water quality.

\subsection{Prediction models and sampling methodology}

Based on different data clustering strategies, three different prediction models were created. One where all measurements were taken as inputs for a single prediction model since all measuring points are geographically near each other. The second and third models were constructed for westernmost $5$ and easternmost $4$ measuring points respectively, where measuring points with similar mean salinity values are grouped. The reasoning behind it is that sources of groundwater considerably influence \textit{E. coli} value, where physical processes for locations with and for locations without those sources can be considered different. If that premise is true, a single prediction model cannot successfully predict for both considered behaviours.

The available data was split into two subsets: 80\% data for training and 20\% for testing. Since it is observed that less than $2\%$ of measurements have contamination levels above the regulation limit, it is expected that random split of training and testing data, such as k-fold analysis, would greatly influence prediction model accuracy, e.g. it is possible that all contaminated measurements end up being sorted in the training set, resulting in high accuracy of prediction model trained on testing set with no measurements above the limit. To take that into account, six different dataset splits of training and testing data were investigated. For the first two datasets, all measurements are sorted by \textit{E. coli} value, and each fifth measurement is taken for the testing set and remaining measurements are used for model training. In this way, the same ratio of measurements above the considered limit is maintained both for the training and testing set. To take into account temporal distribution, two different years are extracted from the dataset so as to serve as test sets, one with smaller and one with a greater number of measurements with reduced water quality where remaining measurements are used as the training set. Similarly, to take into account spatial distribution, two different measuring points are considered for prediction, one with a smaller average \textit{E. coli} value and one with a greater average \textit{E. coli} value. For each prediction model, $20$ runs were conducted to take into account the influence of prediction model parameter randomness and to test the stability of its performance.

\subsection{Cascade prediction model}

\begin{figure}[!ht]
\includegraphics[width=\textwidth]{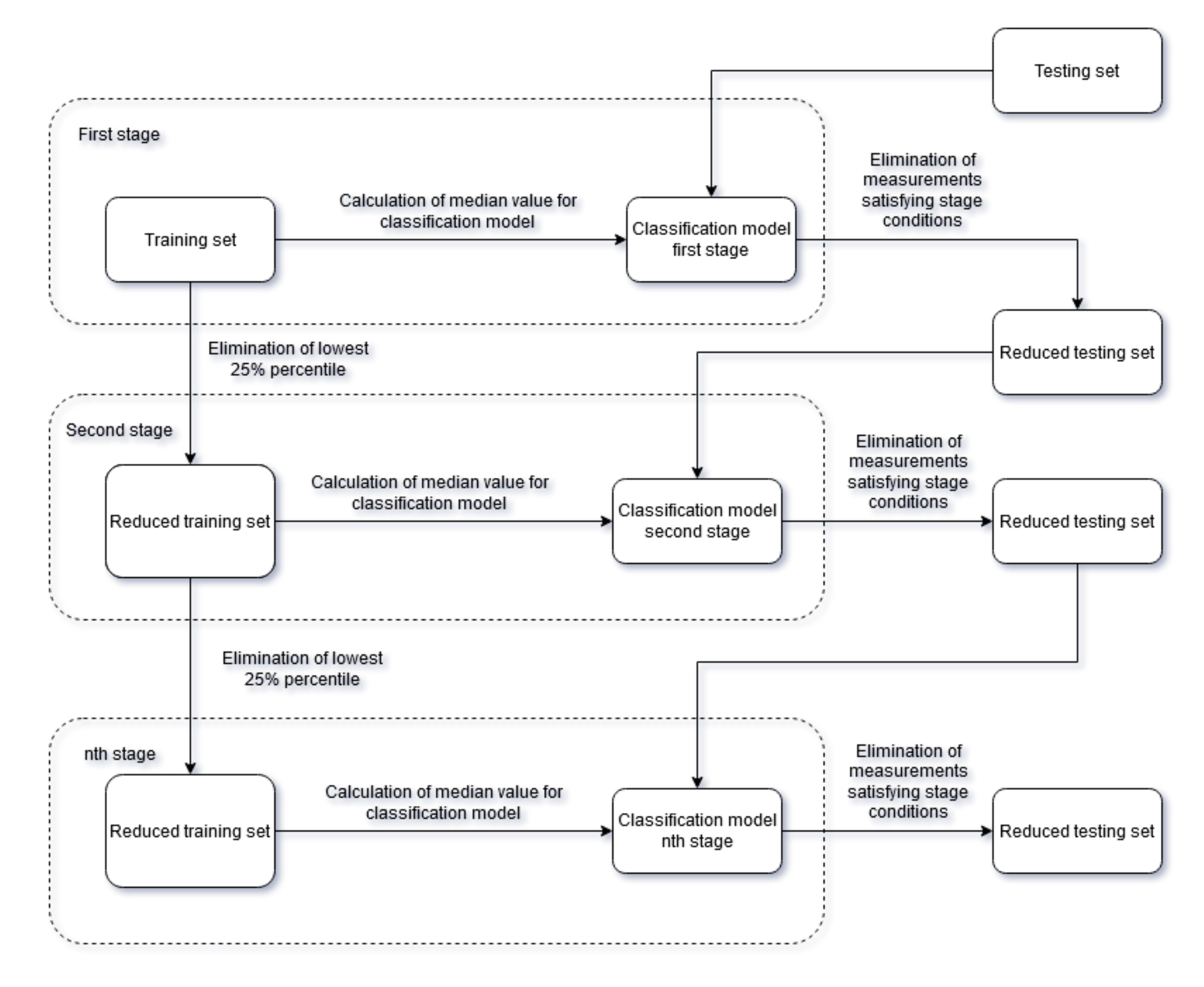}
\caption{Flowchart of proposed cascade model.}
\label{fig:cascade_model}
\end{figure}

Cascade prediction model is considered where classification at every stage is based on the median value of the corresponding dataset which makes the problem fully balanced throughout the cascade. The first stage of the cascade model is trained with the whole training set and the classifier predicts if the measurement is above or below the median of the training set. In the training set of every following stage, the measurements that are below $25$th percentile of \textit{E. coli} value are removed from the training set, resulting in increased median \textit{E. coli} value of the dataset. $25$th percentile and median value were investigated as data reduction limit, however since the reduction of training set size reduces model accuracy, $25$th percentile value is chosen as a good measure. This cascading strategy produces overlapping of training data in several stages, which enables the gradual transition towards greater median values and also a gradual reduction in training set size. Flowchart of the proposed methodology can be observed in Figure \ref{fig:cascade_model}.

With the proposed methodology, the model gradually filters measurements that have low \textit{E. coli} value. If the first stage fails to classify measurement as excellent quality, as a consequence of the gradual transition towards greater \textit{E. coli} values, said measurement can be successfully discarded at the subsequent stage. Additionally, different features' importance is expected, depending on \textit{E. coli} value since the cascade model allows feature weights adjustment at each stage. To improve model reliability, a threshold value for the probability of prediction is introduced. Measurements are considered as excellent quality only if prediction model certainty regarding E.Coli value being below median value is greater than the chosen threshold percentage. Different RF model parameters were investigated and it was observed that change in model parameters greatly influences the efficiency of the threshold approach. This is because model parameters evaluation based on classification prediction accuracy only considers if classification is true or false, and does not consider model certainty regarding prediction, which is the basis for threshold approach utilisation. Ultimately, the best performance was obtained for $800$ estimators, a maximum depth of $10$, and the minimum number of samples required to split an internal node equal to $6$. These parameters are used for all cascade models and for all stages.

Although each stage of the cascade produces imperfect predictions on whether \textit{E. coli} concentration is above or below the median value, these wrong predictions are not problematic for the cascade model as a whole, since ultimately what matters is whether the measurement value is above or below the chosen quality limit. Cascade model was investigated for $250$ CFU/100 mL limit, which is the EU limit for excellent water quality. This results in 63 measurements (5.5\% of all measurements) above the chosen limit, with datasets still having considerable bias.

\section{Results}

\subsection{Random forest classifier - all measuring points}

\begin{figure}[!ht]
\includegraphics[width=0.9\textwidth]{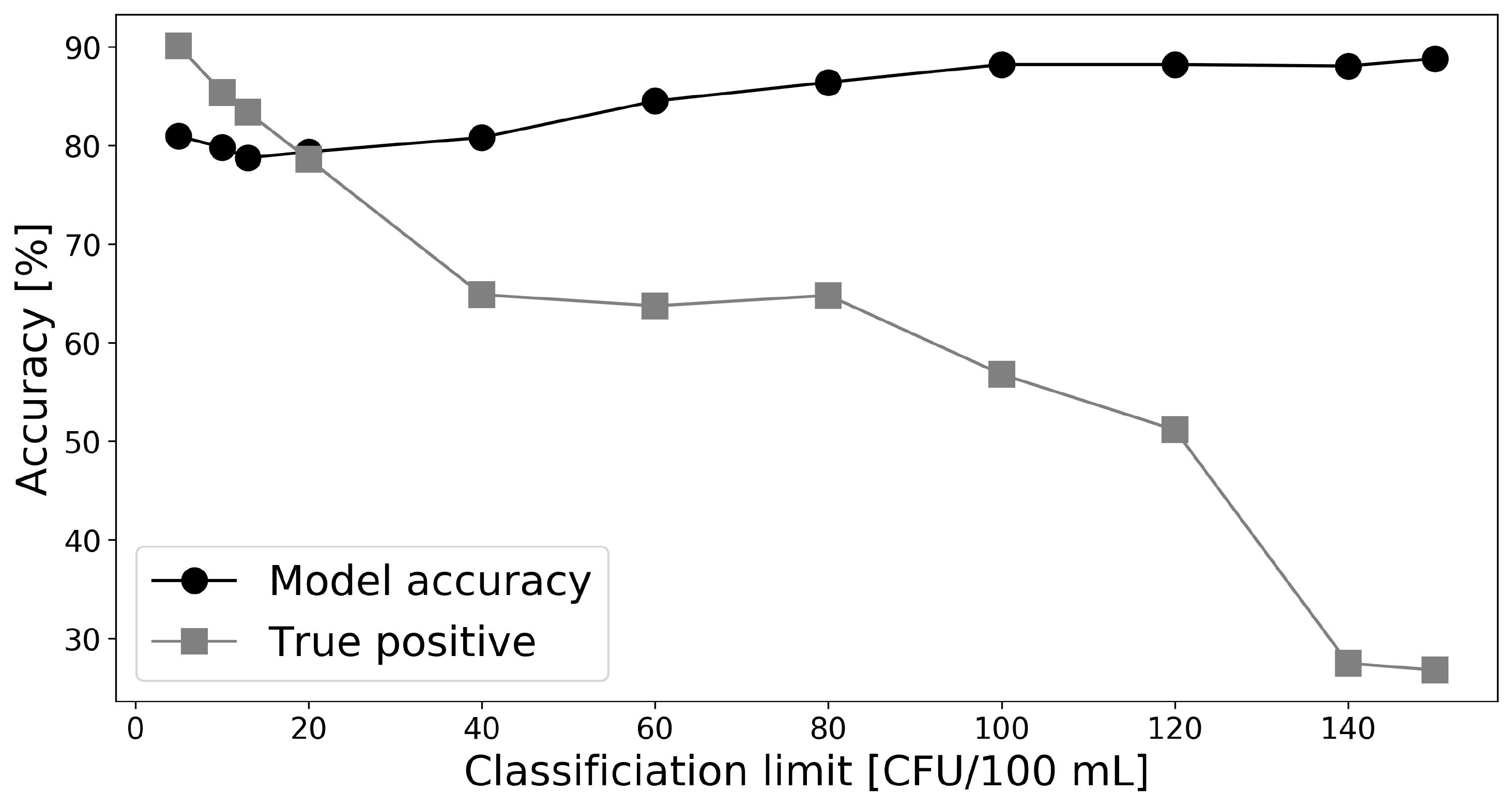}
\caption{Influence of classification limit on prediction model accuracy and true positive rate for Set 1.}
\label{fig:classification_limit_influence}
\end{figure}

Results for the first group of prediction models with all measurement data for different testing-training data splits and for different classification limits are presented in Table \ref{table:prediction_model_testing}. Considered features are water salinity, water temperature, air temperature, GHI, and cumulative GHI for the previous 4 hours which were meteorological data measured during the measurement process with the addition of solar irradiance which is known to positively affect \textit{E. coli} decay \citep{whitman2004solar,berney2006efficacy,maraccini2016solar}. It can be observed that for the EU limit for excellent water quality ($250$ CFU/100 mL) all models have only 20\% of measurements above the given limit correctly classified. That is expected since the number of measurements with reduced water quality is considerably smaller than the number of measurements with excellent quality, thus the prediction model's bias towards excellent quality class yields high model accuracy. When the national limit for excellent water quality ($150$ CFU/100 mL) was used, an increased true positive rate can be observed, albeit it is still very low due to a small number of measurements above the chosen limit. Different classification limits were investigated with the addition of median value to create a balanced split between classes. Results are presented in the Figure \ref{fig:classification_limit_influence}. It was observed that with a lower classification limit problem becomes more balanced and model accuracy increases while model accuracy and true positive rate become similar in proximity to the median limit. To take into account specifics of each training set, it is decided not to consider fixed classification limits, instead, a median value is chosen which always provides a balanced problem through all stages of the cascade model.

\begin{table}[!ht]
\centering
\caption{Prediction model accuracy and true positive rate (TP) for different classification limits of excellent water quality (given in rows) and for different testing sets with indicated number of measurements above considered limit in testing set (given in columns). Results are the average of $20$ runs.}
 \begin{tabular}{ c c c c c c c }
 \hline
{} &\multicolumn{2}{c}{Uniform split} & \multicolumn{2}{c}{Temporal split} & \multicolumn{2}{c}{Spatial split}\\

{} & {Set 1} & {Set 2} & {2019} & {2020} & {KBW} & {KW} \\
{Number of testing} & \multicolumn{6}{c}{} \\
{measurements} & {226} & {226} & {100} & {95} & {144} & {151} \\
\hline
{EU (250 CFU/100 mL)} & \multicolumn{6}{c}{} \\
{Above limit} & {12} & {12} & {16} & {0} & {3}& {12}\\
{Model accuracy} & {94\%} & {95\%} & {83\%} & {/} & {97\%} & {91\%}\\
{TP} & {16\%} & {15\%} & {0\%} & {/} & {0\%} & {15\%}\\
\hline
{CRO (150 CFU/100 mL)} & \multicolumn{6}{c}{} \\
{Above limit} & {23} & {23} & {26} & {3} & {7}& {25}\\
{Accuracy} & {89\%} & {90\%} & {78\%} & {88\%} & {95\%} & {82\%}\\
{TP} & {26\%} & {28\%} & {24\%} & {25\%} & {25\%} & {15\%}\\
\hline
 \end{tabular}
 \label{table:prediction_model_testing}
\end{table}

For each dataset and different classification limits, the analysis of feature importance was conducted. A similar trend was observed for all considered datasets, thus results are reported only for Set1 (Table \ref{table:feature_analysis}). It was observed that for both limits, salinity has the greatest importance, followed by GHI, water temperature air temperature, and cumulative GHI for the last 4 hours. Investigation of other classification limits indicated that the prediction model is more uncertain about its decision about feature importance when a higher classification limit is chosen, which can also be observed here with greater standard deviation for features for EU (250 CFU/100 mL) limit for excellent water quality.

\begin{table}[!ht]
\centering
\caption{Prediction model feature importance for Set1 for different classification limits of excellent water quality. Results are average of $20$ runs and numbers in brackets indicate standard deviation.}
 \begin{tabular}{ c c c }
 \hline
 {} & \multicolumn{2}{c}{Limit} \\
 {Features} & {EU (250 CFU/100 mL)} & {CRO (150 CFU/100 mL)}\\
 \hline
 {Salinity} & {26\% (0.8\%)} & {29\% (0.6\%)}\\
 {GHI} & {23\% (1.1\%)} & {21\% (0.7\%)}\\
 {Cumulative GHI} & {16\% (0.6\%)} & {15\% (0.6\%)}\\
 {Water temp.} & {18\% (0.9\%)} & {19\% (0.6\%)}\\
 {Air temp.} & {17\% (0.8\%)} & {15\% (0.5\%)}\\
 
\hline
 \end{tabular}
 \label{table:feature_analysis}
\end{table}

\subsection{Random forest classifier - separated models}

Further analysis was conducted for two separated models, one with a group of measuring points with higher average \textit{E. coli} value and the second for a group of measuring points with smaller average \textit{E. coli} value. All models were tested on Set1 with uniformly distributed \textit{E. coli} measurements. In order to account for the influence of the training dataset sizes, the datasets for both the model for low \textit{E. coli} values and the model for all measuring points were reduced so as to be of the approximately same size as the model with high \textit{E. coli} values. Each \textit{n}-th measurement is removed from the dataset, so the uniform distribution of \textit{E. coli} measurements is maintained. The results are presented in Table \ref{table:separated_models_same_inputs}. It can be observed that the classification limit has the greatest influence on prediction model accuracy. Grouping of similar measurement locations does not contribute to better prediction accuracy, since for all combinations of grouping and number of inputs, the true positive rate is still below $30$\%. The greatest true positive rate ($82$\%) is achieved for the prediction model with all measurement points and for all available data when the median value is taken as a classification limit. Although this behaviour should be investigated for other training-testing splits, in the subsequent analysis of the cascade model, a single prediction model with all measurement points will be adopted for each stage, as it is evident that the number of measurements, when compared to segregated approach, contributes more to the overall accuracy.

\begin{table}[!ht]
\centering
\caption{Prediction model average accuracy for 20 runs for the three considered models with approximately equal number of inputs and with unreduced dataset. Features considered are salinity, water and air temperature, GHI and cumulative GHI for 4 hours.}
 \begin{tabular}{ c c c c c }
 \hline
 {Low \textit{E. coli}}&\multicolumn{2}{c}{394 train, 99 test} & \multicolumn{2}{c}{526 train, 131 test} \\
 {}&{CRO} & {Median (6.5)} &{CRO} & {Median (6.5)}\\
 {Model acc} & {95\%} & {70\%} & {94\%} & {71\%} \\
 {TP} & {16\%} & {61\%} & {13\%} & {58\%} \\

 \hline
 {High \textit{E. coli}}&\multicolumn{2}{c}{381 train, 95 test} & \multicolumn{2}{c}{} \\
 {}&{CRO} & {Median (38)} &{} & {}\\
 {Model acc} & {83\%} & {75\%} & {} & {} \\
 {TP} & {19\%} & {73\%} & {} & {} \\

 \hline
 {All measuring points}&\multicolumn{2}{c}{383 train, 96 test} & \multicolumn{2}{c}{907 train, 226 test} \\
 {}&{CRO} & {Median (14)} &{CRO} & {Median (13)}\\
 {Model acc} & {79\%} & {79\%} & {89\%} & {78\%} \\
 {TP} & {26\%} & {78\%} & {27\%} & {82\%} \\

\hline
 \end{tabular}
 \label{table:separated_models_same_inputs}
\end{table}

\subsection{Feature analysis}

Further investigation was conducted for rainfall features for the prediction model with all measurement points and median as classification limit. It can be observed from Table \ref{table:prediction_model_median} that the inclusion of considered rainfall features does not contribute to a significant change in prediction model accuracy. It is interesting to observe that the cumulative sum that accounts for rainfall from the 4th to 7th day prior has lower importance than the cumulative sum that accounts for rainfall from the 7th to 14th day prior. Due to having the lowest feature importance, precipitation features were not included in the cascade model testing. However, since it does not reduce model accuracy either, further study should be conducted where inclusion of precipitation features at higher cascade model stages, with greater median values, could be more beneficial.

\begin{table}[!ht]
\centering
\caption{Feature importance for prediction model with all measurement points with median value as classification limit. Results are average of $20$ runs.}
 \begin{tabular}{ c c c c c }

 \hline
{Features} & {Group1} & {Group2}  & {Group3} & {Group4} \\
\hline

 {Salinity}&{38\%} & {41\%} & {40\%} & {43\%} \\
 {Water temperature}&{11\%} & {12\%} & {11\%} & {12.5\%} \\
 {Air temperature}&{10\%} & {11\%} & {10\%} & {11.5\%} \\
 {Cumulative GHI}&{11\%} & {13\%} & {12\%} & {14\%} \\
 {GHI}&{15\%} & {17\%} & {17\%} & {19\%} \\
 {Rainfall 4-7 days}&{6\%} & {7\%} & {/} & {/} \\
 {Rainfall 7-14 days}&{9\%} & {/} & {10\%} & {/} \\
\hline
 {Model acc.}&{81\%} & {80\%} & {81\%} & {80\%} \\
 {TP}&{83\%} & {83\%} & {83\%} & {83\%} \\
 \hline

 \end{tabular}
 \label{table:prediction_model_median}
\end{table}

\subsection{Cascade model results}

The cascade model was first tested on Set1 in order to calibrate the stage parameters. The overview of stage metrics for Set1 can be observed in Table \ref{table:set1_stages_parameters}. Gradual reduction of training inputs and increase of median value through stages can be observed. Six stages were chosen since it was decided that with further stages the size of the training dataset would be too low for prediction model training.

\begin{table}[!ht]
\centering
\caption{Stages parameters for Cascade model for training Set 1.}
 \begin{tabular}{ c c c c c }
 \hline
 {Stage}&{Number of } & \multicolumn{3}{c}{\textit{E. coli} value}\\
 {}&{training inputs} & {median} &  {25\% percentile} & { training range} \\
 \hline
 {1}&{907} & {13} & {3} & {all}\\
 {2}&{690} & {30} & {9} & {$\geq$3}\\
 {3}&{520} & {55} & {20} & {$\geq$9}\\
 {4}&{397} & {80} & {42} & {$\geq$20}\\
 {5}&{298} & {120} & {70} & {$\geq$42}\\
 {6}&{231} & {130} & {92} & {$\geq$70}\\
\hline
 \end{tabular}
 \label{table:set1_stages_parameters}
\end{table}

The cascade model was run $50$ times, where average feature importance through stages is presented in Table \ref{table:set1_stages_features}. It can be observed that at the first stage water salinity has the greatest importance, although as prediction models are trained on measurements with greater \textit{E. coli} value, salinity importance decreases, where other features are given greater importance. This indicates that the introduction of other features at higher level stages could be beneficial for the prediction model.

\begin{table}[!ht]
\centering
\caption{Feature importance for cascade model for training Set 1.}
 \begin{tabular}{ c c c c c c c }
 \hline
 {Features}&\multicolumn{6}{c}{Stage} \\
 {}&{1} & {2} &{3} & {4}& {5} & {6}\\
 \hline
 {Salinity} & {43\%} & {35\%} & {31\%} & {29\%}& {22\%}& {22\%} \\
 {GHI} & {18\%} & {18\%} & {20\%} & {23\%}& {25\%}& {27\%} \\
 {Cumulative GHI} & {14\%} & {18\%} & {18\%} & {18\%}& {17\%} & {17\%}  \\
 {Water temperature} & {12\%} & {14\%} & {16\%} & {18\%} & {19\%} & {20\%} \\
 {Air temperature} & {12\%} & {14\%} & {15\%} & {13\%} & {17\%} & {15\%} \\

\hline
 \end{tabular}
 \label{table:set1_stages_features}
\end{table}

The influence of threshold value on model accuracy results can be observed in Table \ref{table:set1_threshold}. The same threshold value is set for all stages. Greater number of measurements are predicted as excellent quality with a lower threshold value, albeit with a greater percentage of wrong predictions. For most datasets, the threshold value of $80$\% assures there is no elimination of days with \textit{E. coli} value greater than $250$ CFU/100 mL, with the exception of a single measurement for the $KW$ dataset which is still incorrectly predicted even with a threshold of $85$\%. This measurement could be an outlier or additional cascade model improvement could eliminate this wrong prediction. Thus, for the purpose of further study, the threshold value of $80$\% is adopted.

\begin{table}[!ht]
\centering
\caption{Influence of threshold value on cascade model accuracy for different datasets. Numbers in brackets next to datasets indicate total number of measurements in testing set and number of \textit{E. coli} measurements above EU quality limit. Presented results are average of 50 runs.}
 \begin{tabular}{ c c c c c }
 \hline
 {Set1 (226, 15)}&\multicolumn{4}{c}{Threshold} \\
 {Excellent quality prediction}&{85\%} & {80\%} &{75\%} & {70\%}\\
 \hline
 {True positive} & {72 (34\%)} & {91 (43\%)} & {106 (50\%)} & {125 (60\%)}\\
 {False negative} & {0 (0\%)} & {0 (0\%)} & {0 (0\%)} & {0 (0\%)} \\

\hline
 {Set2 (226, 15)}&\multicolumn{4}{c}{} \\
  \hline
 {True positive} & {65 (31\%)} & {96 (45\%)} & {117 (55\%)} & {140 (66\%)} \\
 {False negative} & {0 (0\%)} & {0 (0\%)} & {0.28 (2\%)} & {1 (7\%)} \\

  \hline
   {KBW (144, 3)}&\multicolumn{4}{c}{} \\
  \hline
 {True positive} & {82 (58\%)} & {84 (59\%)} & {96 (68\%)} & {105 (74\%)} \\
 {False negative} & {0 (0\%)} & {0 (0\%)} & {0 (0\%)} & {0 (0\%)}  \\
  \hline
     {KW (151,12)}&\multicolumn{4}{c}{} \\
  \hline
 {True positive} & {34 (24\%)} & {55 (39\%)} & {71 (51\%)} & {84 (61\%)} \\
 {False negative} & {1 (8\%)} & {1 (8\%)} & {1.7 (14\%)} & {3 (25\%)}  \\
  \hline
 {2019 (100,16)}&\multicolumn{4}{c}{} \\
  \hline
 {True positive} & {4 (5\%)} & {11 (13\%)} & {27 (32\%)} & {41 (48\%)} \\
 {False negative} & {0 (0\%)} & {0 (0\%)} & {1 (6\%)} & {2.3 (14\%)}  \\
  \hline
 {2020 (95, 0)}&\multicolumn{4}{c}{} \\
  \hline
 {True positive} & {34 (36\%)} & {50 (52\%)} & {59 (62\%)} & {70 (74\%)} \\
 {False negative} & {0 (0\%)} & {0 (0\%)} & {0 (0\%)} & {0 (0\%)}  \\

  \hline
 \end{tabular}
 \label{table:set1_threshold}
\end{table}

To further enhance the proposed model, a combination of threshold values for different stages was analyzed. Since in the first few stages the median value is considerably low, it is reasonable to assume that a smaller threshold value in those stages could still provide good results. For the first stage threshold value of $65$\% is considered, for the second stage $70$\%, for the third stage $75$\%, and for all other stages $80$\%. Secondly, due to the overlapping of training data through stages, an additional check is introduced: if both in current and previous stage prediction model certainty is above the chosen threshold, which is chosen to be lower than the threshold for the current stage, then measurement can also be classified as below limit. This can be understood as that two weak certainties at subsequent stages can be considered as one strong certainty in the current stage. The considered threshold values were $70$\% for the second and third stage and $75$\% for the remaining stages. Additionally, the influence of different features throughout the stages was also considered. Due to its small feature importance in the first three stages, the air temperature was removed as a feature and then introduced only in the last three stages. Combinations of these methods were also investigated. 

Overview of the obtained results can be seen in Table \ref{table:set1_stage_adjustment}. It can be observed that all proposed methods increase the number of excellent quality predictions, where the combination of all three methods further increases that number. It must be noted that only one combination for each method was presented, where further investigation of more combinations could provide better results, e.g. different features, different values of increasing threshold value, etc.

\begin{table}[!ht]
\centering
\caption{Influence of stage adjustment on cascade model for Set1. Presented results are average of 50 runs with no wrong predictions of excellent water quality.}
 \begin{tabular}{ c c c c c}
 \hline

 {}&{Remaining measurements} & {Right prediction (\%)}\\
 \hline
  {Feature change} & {136} & {43\%} \\
   {Additional deletion} & {125} & {48\%} \\
 {Increasing threshold (65-80\%)} & {116} & {52\%} \\
 {All} & {113} & {54\%}\\
\hline
 \end{tabular}
 \label{table:set1_stage_adjustment}
\end{table}

Proposed improvement of the cascade model with all three adjustments was tested on different datasets and results are presented in Table \ref{table:cascade_different_sets_all3}. It can be observed that the number of correct predictions of excellent quality is greatly influenced by the testing set. Also, for some datasets wrong predictions are also observed, indicating that more rigorous adjustment of proposed cascade model improvements should be conducted.

\begin{table}[!ht]
\centering
\caption{Cascade model results for prediction of excellent water quality with all 3 methods for different datasets.}
 \begin{tabular}{ c c c c c}
 \hline
 {Dataset}& {Number of}& {Measurements with}& {True}& {False}\\
  {}& {measurements}& {reduced quality}& {positive}& {negative}\\
  \hline
{2019} & {100} & {16}& {27 (32\%)} & {0.02 (0.13\%)} \\
{2020} & {95} & {0}& {61 (65\%)} & {0 (0\%)} \\
{KBW} & {144} & {3}& {71 (51\%)} & {0 (0\%)} \\
{KW} & {151} & {3}& {71 (51\%)} & {0 (0\%)} \\
{Set1} & {226} & {15}& {113 (54\%)} & {0 (0\%)} \\
{Set2} & {226} & {15}& {118 (56\%)} & {1.24 (8\%)} \\
\hline
 \end{tabular}
 \label{table:cascade_different_sets_all3}
\end{table}

\section{Discussion}

\subsection{Prediction modeling}

From the conducted analysis it can be concluded that a single prediction model for prediction of \textit{E. coli} value being above or under limits $150$ or $250$ CFU/100 mL does not provide satisfactory results due to considerable data bias. The prediction model tends to classify all measurements as excellent quality since they make a vast majority of the measurements. Thus, a cascade model is introduced which is shown to work well as a filter of measurements with excellent quality, without removing measurements above the chosen limit. The proposed cascade model has balanced datasets at each stage since the median value is considered as a classification limit at each stage. However, different limits can also be explored to possibly further improve model accuracy. It must be noted that in general, measurements considered in this paper have overall excellent water quality, however, the presented methodology can be used for different pollution values since the cascade model is constructed to adapt to the provided data.

It was observed that the salinity feature has the greatest importance in single prediction models, and also for several first stages of the cascade model. However, in further stages, it was observed that salinity value is not as important. It indicates that the salinity feature has the greatest importance for predicting measurements with low \textit{E. coli} value. Since the majority of measurements for the single prediction model have low \textit{E. coli} value, a strong weight is put on the salinity value which is beneficial for the majority of data but is not beneficial for the right prediction of high \textit{E. coli} measurements, which are in fact most important. Thus, the cascade model enables adjustment of feature importance through stages for different levels of pollution. Additionally, some features that could be important for the right prediction of higher \textit{E. coli} values would decrease accuracy for low \textit{E. coli} values, thus different features can be included at different stages. In order to maximize the capabilities of the cascade model approach, precise feature engineering should be conducted.

It must be noted that RF model parameters were the same for all stages and proposed improvements for the cascade model were investigated only for one combination of parameters. Further investigation of model parameters through stages and further study of different combinations of proposed improvements should be conducted to further increase cascade model efficiency. Currently, both cascade and single prediction models are constructed with measurements from multiple measurement points due to the small amount of data. However, ultimately separated prediction models could be constructed to establish a relation between \textit{E. coli} value and specific processes for that measurement point.

\subsection{Data analysis}

From in-depth analysis of measurement data, it was observed that sources of groundwater are related with the greater value of \textit{E. coli}. This was also corroborated by the higher importance of salinity in the first couple of stages of the cascade model. High salinity value corresponds to summer months with longer dry periods, where \textit{E. coli} values are very low (as are classification limits for the first several stages), where the prediction model heavily relies on that information to consider a measurement as excellent quality. As salinity value decreases, it most often corresponds to spring months where the sea is still influenced by longer periods of rainfall, or alternatively to measurements influenced by coastal springs. For both of these occurrences, it is important to consider solar irradiation and both water and air temperature, which is supported with more evenly distributed feature importance at higher stages.

\begin{figure}[!ht]
\includegraphics[width=\textwidth]{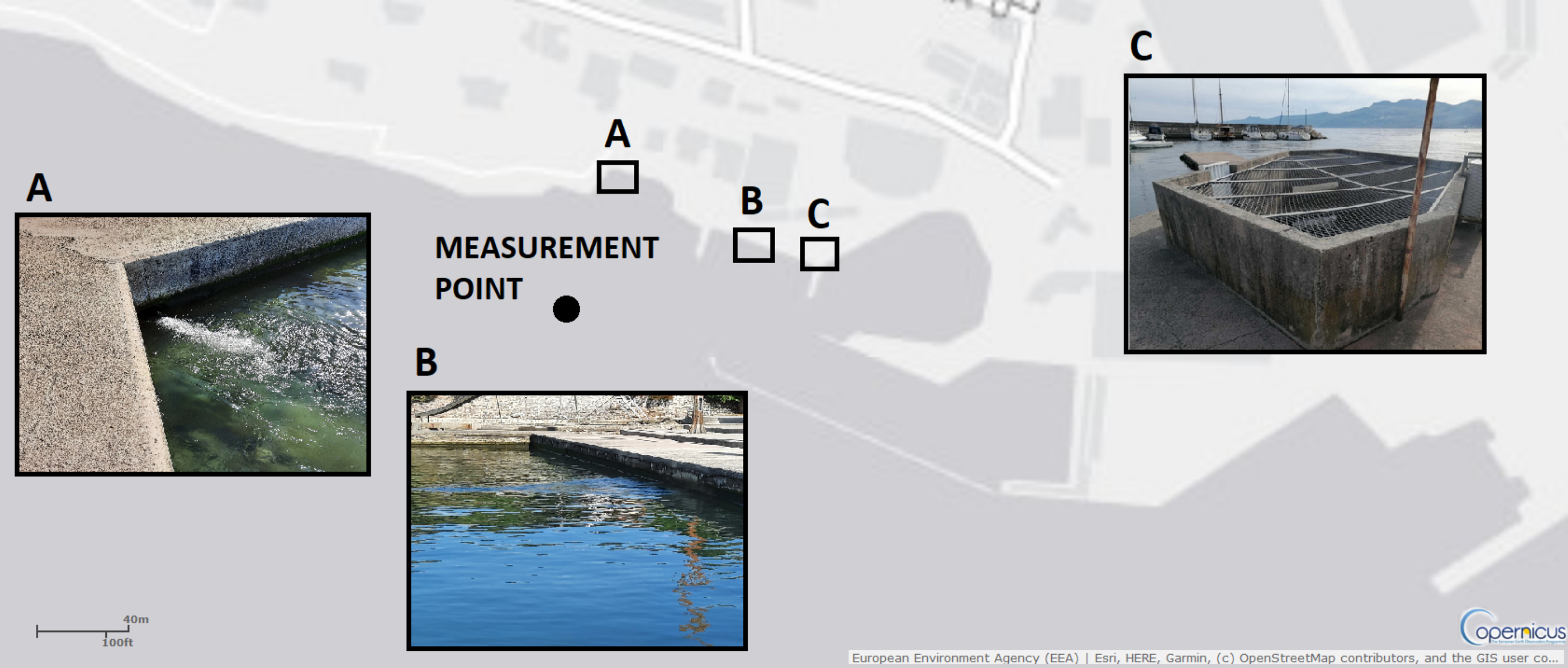}
\caption{Detail of 3M beach location with indicated some of the springs locations. Location C is collector of multiple springs. }
\label{fig:springs_locations}
\end{figure}

Groundwater springs are of great interest when they can be used as a source of drinking water. Since coastal springs that are of interest in this study are brackish springs, smaller and with seasonal character there is no study of their characteristics in the literature. Additional in-field investigation of 3M location was conducted during May 2021 after a longer period of rain where a considerable number of submerged spring sources were observed (Figure \ref{fig:springs_locations}). Although the occasional presence of springs is well known since they can be experienced as cold areas during swimming, they were not previously considered as being of strong importance and were thought to dry up during summer. However, observation of these springs is mentioned in \cite{stravzivcic1999rijevcki}, where two big springs which are active throughout the whole year are said to exist at the 3M location. Furthermore, another three springs are mentioned, where the eastern spring, which is closest to the bathing location, is active yearlong and the other two usually become inactive only by the end of summer. In KE location one strong spring is mentioned with the addition of the number of smaller springs throughout the coast.

\begin{figure}[!ht]
\includegraphics[width=\textwidth]{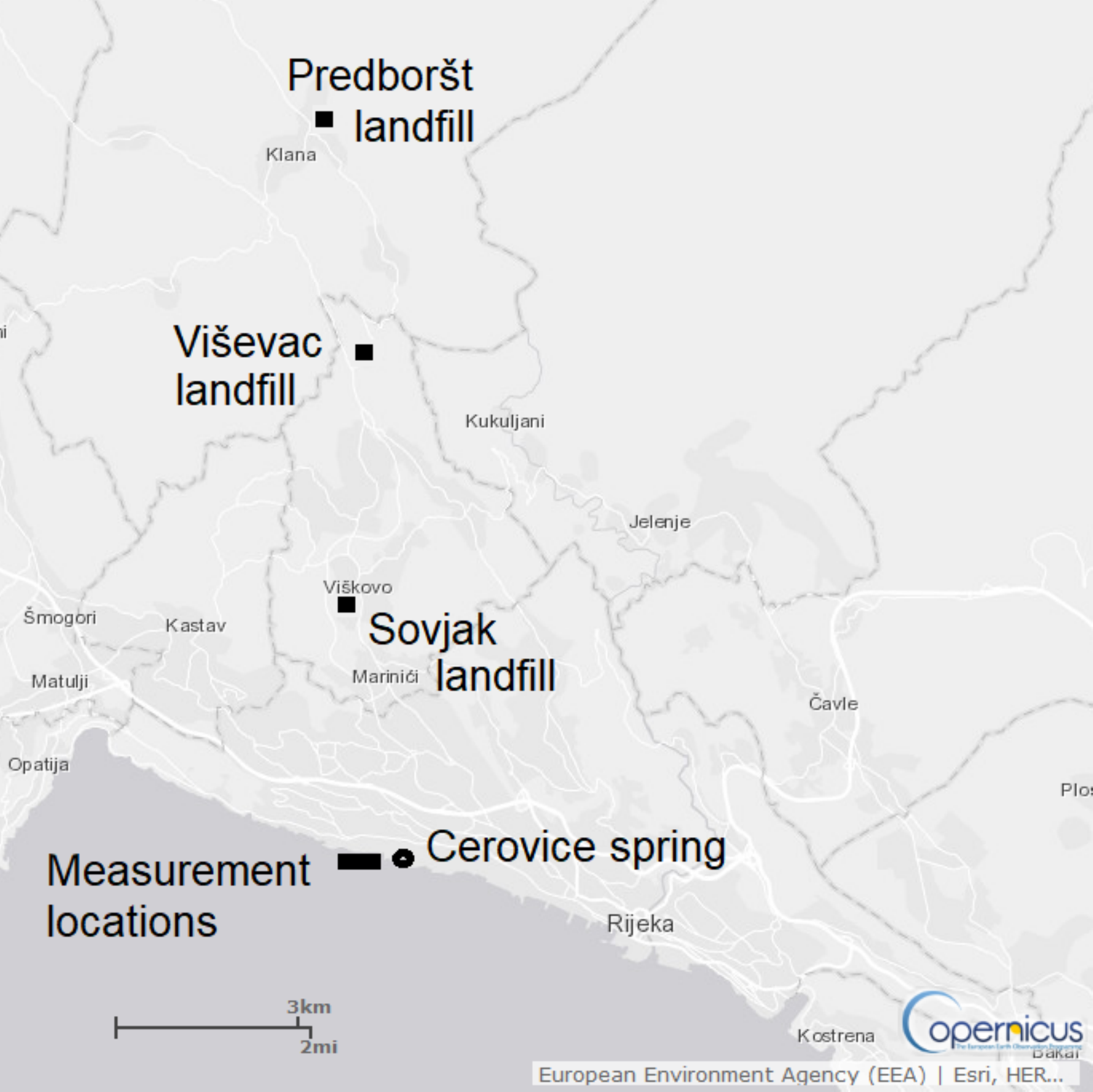}
\caption{Locations of spring Cerovice and landfills in the hinterland of measurement locations. }
\label{fig:landfills_locations}
\end{figure}

Additionally, in the immediate vicinity of the beach 3M is greater spring Cerovice (Figure \ref{fig:landfills_locations}) which is located inside shipyard "3. Maj". It consists of several springs that are active throughout the whole year, but since microbiological pollution was observed that water is only used as technical water. This could explain reduced water quality for measurement locations 3M and KE, which are specific since they have a considerable amount of springs active yearlong, probably with water quality similar to spring Cerovice. Reduced water quality in KW location could be due to the influence of KE springs since wind and sea currents can cause transport of contamination. Additionally, in the study by \citet{biondic2009ocjena} risk assesment of underground water was conducted, where it was indicated that landfills in the hinterland of city Rijeka can influence coastal springs in the area considered in this study (Figure \ref{fig:landfills_locations}). 

These observations indicate that current data, comprising both from regular measurements taken from the boat (approximately 20 m from the coastline) and additional measurements taken from the coast, could give considerably different results under the same meteorological conditions if additional sampling is conducted in the immediate vicinity of the spring location. Both types of measurements were included to increase the teaching dataset which should lead to improved model accuracy, however with these new findings, this could in fact reduce prediction model accuracy. Thus, several directions of further research are suggested. 

\subsection{Further research}

First, analysis of the coastal area should be conducted in order to investigate the number and locations of possible coastal springs. It is known that some sources are active only for some periods, where some sources are active almost all year. Thus, an investigation of the source activation period should be conducted. Observation of these sources presented in \cite{stravzivcic1999rijevcki} only notes their existence and locations while activation periods and discharge values should be also known so as to assess their influence on bathing locations. Also, these observations should be revised and adjusted, due to changes in city infrastructure (such as drainage and sewage systems reconstructions) conducted in the period of the last 20 years or more. Additionally, every beach location has its specific geomorphology, where the number and size of springs are different, as well as its openness to the sea. In this study, 3M location was further inspected (Figure \ref{fig:springs_locations}) which is a closed, small port, and as such under a strong influence of the observed springs. Other locations, which are less enclosed, are expected to show a smaller influence of springs, however, this premise should be further investigated.

The second direction of future research should investigate rainfall influence and watershed area more extensively. Different rainfall correlations with increased microbial pollution are found through literature, e.g. previous 24 hours \citep{mallin2001demographic}, 48 hours \citep{kelsey2004using}, 7 days \citep{lipp2001effects}, 5.5-9.5 days \citep{de2018optimising} and even 30 days \citep{pandey2012assessing}. This is a strong indication that geological specifics must be taken into consideration. Precipitation considered in this work was only from a single location for the city center of Rijeka, which can influence short-term water influx to the sea. However, during summer periods it can be expected that dry ground absorbs a great amount of rainfall. This could explain why there was no considerable correlation with cumulative sums of rainfall from several previous days in the city center. Additionally, karst groundwater is greatly influenced by precipitation from the whole watershed area. For example, tracer tests were conducted in \citet{biondic3slovenski} where the tracer was injected in V. Sne\v{z}nik (Slovenia) where underground water connection with Kvarner Bay area was identified, including measurement locations considered in this study, indicating transboundary characteristics of the investigated area. Thus, further research should include an investigation of correlations between coastal springs' activation and wider regional area's precipitation, as these distant rainfalls are expected to possibly be more influential for coastal spring activity than local rainfall. It is also important to mention that a boundary between two watersheds is passing through the center of the city of Rijeka, thus bathing locations on the east side of the city are expected to show a different behaviour, and different locations for rainfall measurements should be considered.

Ultimately, the measurement process of \textit{E. coli} could be improved on the grounds of these new observations. If a stronger influence of groundwater sources is observed for some bathing locations, and if those sources are observed to often have reduced water quality, a unique measurement methodology should be established for these locations, where multiple locations, contrary to the current single measurement location, should be investigated to give a better description of the considered bathing location. Additionally, in \citet{rukavina4vavznost} the bathing locations in the east part of Rijeka were analysed, and springs were found to have unusually and considerably greater value of \textit{E. coli} then values observed at the monitoring location. Follow-up investigation led to the identification of one business located approximately 2 km upstream, with inappropriate wastewater connection as a source of pollution which was connected with observed springs. This indicates that springs are a great vulnerability of bathing locations, especially if they have a greater outflow and are influenced by greater watershed area where periodic monitoring of such springs should also be considered.

\section{Conclusion}

In the presented paper, in-depth data analysis of \textit{E. coli} measurements and related data was conducted and a predictive machine learning modeling strategy was proposed. Due to a very small amount of measurements with reduced water quality in the database, it was shown that a single prediction model has reduced prediction accuracy due to bias toward classifying all measurements as excellent quality. Thus, a cascade model approach is proposed which classifies measurement as excellent quality only if it is highly certain regarding its decision. Other measurements remain suspect, therefore the proposed method can be considered as a filter method, which can be further improved to be used as a standalone model. The following observations regarding the prediction model can be made:
\begin{itemize}
    \item {Grouping of measurement points does not provide improvement in prediction model accuracy since an increased number of inputs is more beneficial.}
    \item{Due to bias of input data it is difficult for a single prediction model to confidently predict occurrences of subpar bathing water quality.}
    \item{Cascade model can provide a predictive data filter in which excellent water quality can be predicted with high accuracy, based on meteorological data, solar irradiation, and seawater salinity.}
    \item{Due to the high flexibility of cascade model, multiple strategies of its improvement could possibly lead to it being eventually used as a standalone prediction model.}
    \item{Ultimately, a separate prediction model for each measurement point could be constructed, to capture the uniqueness of each beach, which is especially important in the karst type of terrain.}
\end{itemize}

Conducted study indicates a strong need for an interdisciplinary approach for the given problem. The karst type of soil with its specific underground landscape absorbs rainfall from a large watershed area, indicating not only rainfall period but also rainfall measurement locations are important. Additionally, \textit{E. coli} measurements could be conducted for the number of known springs to indicate which springs contribute to the coastal seawater pollution and in which amount. Regarding \textit{E. coli} measurements, the following is observed:
\begin{itemize}
    \item {Submerged groundwater sources, active yearlong, are observed which could correlate with greater \textit{E. coli} value.}
    \item{Number of groundwater sources and their intensity should be investigated at the considered bathing locations.}
    \item{New sampling methodology should be established to take into consideration the mixing of microbiologically contaminated submerged spring water and seawater.}

\end{itemize}

\section{Acknowledgements}

This research article is a part of the project \textit{Computational fluid flow, flooding, and pollution propagation modeling in rivers and coastal marine waters - KLIMOD} (grant no. KK.05.1.1.020017), and is funded by the Ministry of Environment and Energy of the Republic of Croatia and the European structural and investment funds. Also, the authors acknowledge the funding of the University of Rijeka through the project \textit{Razvoj hibridnog 2D/3D modela za u\v{c}inkovito modeliranje strujanja u rijekama, jezerima i morima}. Furthermore, authors acknowledge the support of the Center of Advanced Computing and Modelling at the University of Rijeka for providing computing resources.

\bibliographystyle{elsarticle-harv} 
\bibliography{cas-refs}






\end{document}